# Models of robotic feeding, choice, and the survival mechanism

*Christopher A. Tucker*
*Cartheur Robotics*

cartheur@gmail.com

Abstract

Diagrammatic models of feeding choices reveal fundamental robotic behaviors. Successful choices are reinforced by positive feedback, while unsuccessful ones by negative feedback. This paper will address robotic feeding by casually relating consequential behavior subtended by a strong dependence upon survival.

**Introduction**

Recharging models for an autonomous robotic platform has been classically viewed in the context of behavioral responses, yielding interesting insights [1–4]. A central theme is the consideration of consequential behavior to the design of a recharging model to allow freedom of emergence [5–8]. Given a choice, this freedom is enforced. This paper will describe generalized recharging models and discuss the behavior exhibited by a robot pursuing a source of food, i.e., replenishing the energy in its onboard systems. Concepts such as transformation between states and feedback will be introduced as the mechanisms of choice, and a novel feeding model will be proposed which provides an additional freedom of expression of the survival instinct.

W. Grey Walter first proposed feeding behavior in artificial systems as a means to understand fundamental characteristic behavior in living systems [1, 5, 6]. Owen Holland notes [9]:

> In his writings about the tortoises, Grey Walter gave much weight to an attribute he called 'internal stability'—the claimed ability of the tortoises to maintain their battery charge within limits by recharging themselves when necessary. A feature of the tortoises' circuitry was that, as the batteries became exhausted, the amplifier gain decreased, making it increasingly difficult to produce behavior pattern N (negative phototropism).

By purposefully including circuital features to manipulate responses for an activity such as recharging, Walter was able to foster emergent behavior in his tortoises. He suggested that an artificial system could be designed in such a manner to study behaviors commonly witnessed in biological systems [10]. In light of Walter's work, this paper proposes two research questions:

1. Can a robot be given the ability to make its own choice?
2. Can a robot be made aware of a dependency between its choices and survival?

To facilitate answers to these questions, four goal-based assumptions [11, 12] illustrate the necessary parameters in the model:

1. The artificial system under examination is fully autonomous, that is, once the system is started it requires no further input from an operator. In order to be autonomous, the system is self-sufficient, e.g., it has the necessary components for its operation and runs continuously.
2. The artificial system exists in situ with its environment and composes algorithms in response to its interaction with it.
3. The artificial system possesses a system of behaviors relevant to its purpose, the ability to evolve, and a set of choices within the scope of its design.
4. The artificial system leverages behaviors indistinguishable from biological systems, from the observer's point of view. To be an effective model, a principle of equivalence illustrates the behaviors are archetypical, e.g., such behaviors are essentially identical for a biological organism with similar environmental pressures.

This paper will present a characteristic model of robotic feeding. It will address the model first in a generalized form, increase its complexity, then introduce a wireless-power delivery method containing a more colorful set of feeding behaviors. Lastly, it will describe a novel circuit that introduces the capability to ascribe the survival instinct to the robot. It is the goal of this paper to illustrate a method and means of the quantification in an algorithm of the consequence of choice and decision-making. It is argued such a study is valuable not only for artificial but for biological systems as well to better understand primal features of life and that they are far more accessible than once believed. Such an understanding evolves designs of robotic components that mimic natural forms giving them greater independence in environments undergoing changing conditions. It is impact of the realization that a richer set of behaviors offers the capability for study of the phenomenon of the survival instinct not only artificial systems but also for living systems.

**General model of robotic feeding and the application of choice**

Apart from strict considerations as a form of robust control [13] or event-driven agents [14], robotic feeding considered here is analogous in form and function to an activity [15] exhibited by organic entities, and can be reduced to a simple model of goal-seeking behavior. Such a model is illustrated in Fig.1.

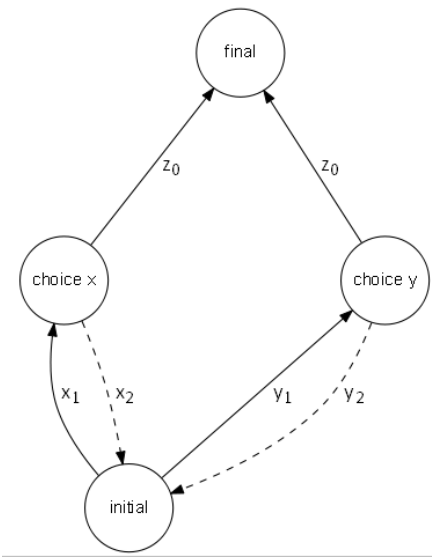

Fig.1. Activity of robotic feeding behavior.

A robot that is seeking power to recharge its onboard power system begins its activity at *initial* and is presented with one or more choices—in this example, two choices labeled $x$ and $y$—whereby to reach its necessary goal *final*. In order to decide which path to pursue, $x_1$ and $x_2$ toward $x$, $y_1$ and $y_2$ toward $y$, weights are assigned based on either success or failure of the path leading to the pursuit of *final* at $z_0$. Through repetition of this activity of seeking power, consecutive weights are averaged and the robot "prefers" pursuit of one path over the other because of positive experiences as well as negative feedback. Pseudocode of this activity is shown in Fig.2.

```
Activity of robotic feeding behavior - Pseudocode

    --Task--

    Recharge the batteries present in the system before power is exhausted.

    --Activity--

    Notice that the power level to sustain continuous operation is low enough to require recharge.
    Search stored data for available recharging types, of these types retrieve the weighted values
    to determine which is most optimal. If these values are equal to zero, generate a random number
    to choose which choice to pursue.

    Pursue choice 'x'. If recharging is reached, store a value of one for variable 'x1'; if
    recharging is not reached, store a value of zero for variable 'x1'.

    If recharging is not reached from pursuit of choice 'x', pursue choice 'y'. If recharging
    is reached, store a value of one for variable 'y1'; if recharging is not reached, store a
    value of zero for variable 'y1'.

    When the task is called, collect the weighted values for each recharging type. Sort the values
    in descending order. Pursue those choices on the list which have greater values. When pursing
    a choice by greater value and recharging is not reached, divide the value by two and store
    the new value.

    --End Activity--
```

Fig.2. Activity of robotic feeding behavior – Pseudocode.

The notion of the activity as a template for robotic feeding behavior serves as the primary theme for the description of the environment containing the robot. As such, the template can be expanded to include more detail relevant to ascribed behavior. In terms of defining a set of environmental factors which facilitate the activity of power-seeking behavior, the process is modeled as a run-to-completion state machine, shown in Fig.3.

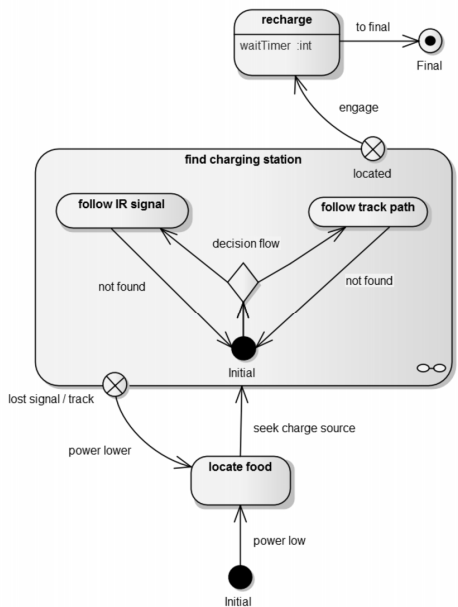

Fig.3. Finite-state diagram of robotic power-seeking behavior.

The components of Fig.3 depict the template in terms of a typical finite-state machine for a robot tasked with finding a recharging source. In ascribing behavior in an empathetic context, it is performing the task of searching for food. This activity is started when notified by the event *power low* wherein it will *locate food*. It will execute *seek charge source* entering the state machine.

The states, represented as boxes, are: *Initial*, *locate food*, *feeding*, and the state machine *find charging station* that contains *follow IR signal* and *follow track path*. The actions, represented as crossed circles, are two exits from the state machine. One is for a positive result, *located*, and one is for a negative result, *lost signal/track*. The transformations, represented as arrows: *power low*, *power lower*, *located*, are consequences of the choice following *seek charge source*. The transformation at the junction of *decision flow* indicates the decision since more than one outcome is present and the choice is made consequential of environmental factors. The software controlling the decision stipulates, without optimization, that it based on positive sensor feedback—if the IR signal or the track path is discovered first. The first acted upon, the alternative discarded unless the former returns a negative result.

The robot enters the state machine at the *Initial* orb when the sensor responsible for monitoring battery level notifies the operating system that power is low, noted in the transformation. When within the action *locate food*, a routine in the program executes the behavior for optimal seeking of a charge source. The transition of this behavior leads to entry into the *find charging station* state machine at *Initial*. If a positive result is obtained—that either of the choices are successful—the robot exits at *located*, and the transformation *engage* leads to the state *recharge*. When *waitTimer* expires, it will exit at *Final*. In terms of the complete behavior in this diagram, most of the complex behaviors are executed in the state machine, given the choice in the decision flow between to follow an infrared (IR) signal or follow a track path. Existence of such a choice is highly dependent on multiple solutions to the charging problem, if the state machine did not have both an IR source and track path to power to guide the robot, then choice in this context is irrelevant. In Fig.3, the experience derived from results of trying to follow one branch or another—found or not found—is one case of behavior. The experience derived from the pursuit of the specific choice—*located* or *lost signal/track*—is a second case. In the first case, not finding an IR source or a track path could be the result of neither existing nor unable to be found due to the causality of a sensor function designed to detect them. In the second case, having found the IR source or the track path but not locating it will keep motivation to continue finding it, or the robot remaining inside *find charging station*. When the state machine fails to return a positive result, it will exit at *lost signal/track*, the transformation then notes *power lower*, when compared to the transformation *power low*.

*Decision-making embedded within transformation logic*

Choice, in the scope detailed here, is a phenomenon isolated in the transformation between states yielding a consequence of one outcome. Given the power to select one outcome from many, the weight of consequence becomes determinate, e.g., one decision more optimal than another, to within a tolerance of 0.1 between weight values. From Fig.3, once the event for *locate food* is triggered, Fig.4 represents the behavior in the state machine *find charging station*.

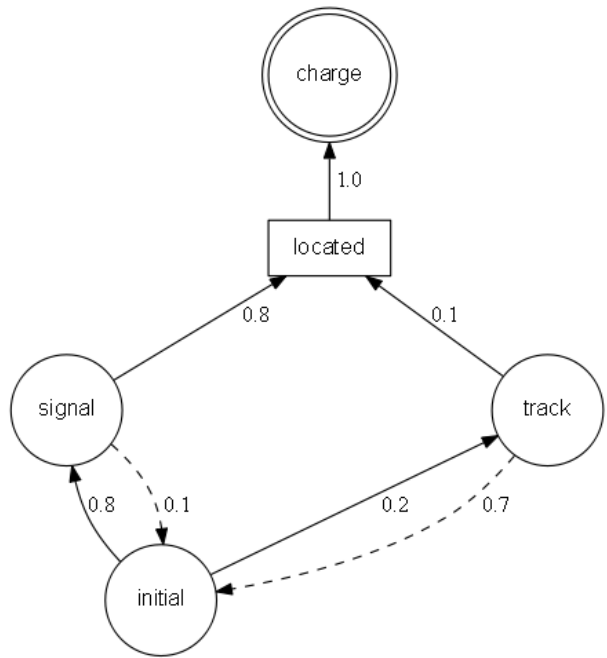

Fig.4. Runtime choice-weights for consequential decision-making.

The robot enters the diagram at *initial* and by reading the weights, can determine that recharging by using *signal* is better than *track* given the comparison positive weights (solid line) are 0.8 and 0.2, the comparison negative weights (dashed line) are 0.1 and 0.7, respectively. Within the context of the program, the weights for each decision path are averaged for each successful result. Each time the robot enters the *find charging station* state machine, it will learn to choose the optimal path because of the higher value of the weight. If decision paths have the same weight, a choice that is sufficiently random would assign a decision. The ethological implication of the modeled behavior embedded in the diagram of Fig.4 is the dynamic of it at different points of time during the activity of seeking a feeding source. The term "feeding" is applied here in the same scope as its original biological conception, that an entity pursuing food—in the case of the robot, energy—is exercising an adaptation for optimization of its survival.

According to Ashby [16, 17], states and their transformations are constructs of a characteristic map of behavior leading to the thinking process of entities, as noted here by the weights for each decision including positive and negative feedback. The implication is the fitness of the model and its completeness. What is illustrated in the runtime diagram are the degrees of change that the robot goes through during a finite quantity of time while attaining its goal. The model does not try to reveal the mechanisms behind the operations directly, rather, the character of the transition between states alluding to the behavior of the sequence. The goal is to reveal behavior of the robot during its power-seeking activity and gain evidence for the survival instinct in artificial systems. To accomplish this goal, in addition to a standard battery-charging station, a wireless charging model in presented next.

**A wireless-power model of robotic feeding**

In the previous model, a greater magnitude of freedom is offered by the presence of an IR signal to help guide the robot to its feeding source—as opposed to only having a track path. The presence of the signal allows the robot to receive information about the source, finding it more readily and efficiently, as illustrated by the weights in Fig.4. The wireless-power feeding model builds upon this, adding more degrees of freedom to the activity. The combined model in the form of a state machine is shown in Fig.5.

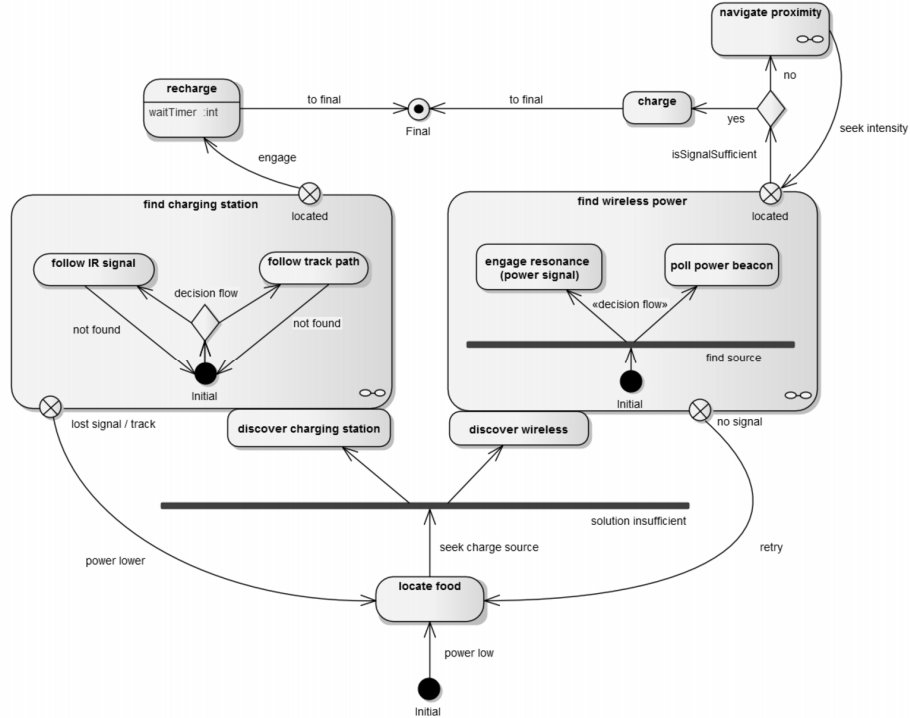

Fig.5. Finite-state diagram of power-seeking behavior for charging station and wireless power.

The components of Fig.5 encapsulate behaviors in the same manner as Fig.3, however, seeking has an additional dimension associated with it: there is also the presence of a power signal that only requires a decision to engage a connection to feed, allowing greater cooperation between the robot and its environment [18]. To the left of Fig.5 is the *find charging station* state machine with its exit *located* leading to the transformation *engage* where it will stay in *recharge* until *waitTimer* determines when recharging is complete, wherein it leaves at *Final*. To the right of Fig.5 is the *find wireless power* state machine with its exit *located* leading to the Boolean decision of *isSignalSufficient* will tell the robot if it is in a suitable proximity to the power signal to perform recharging at *charge*, else to *navigate proximity* where it will *seek intensity* until the Boolean is satisfied. If true, it leaves at *Final*.

In the method demonstrated by Fig.5, when the robot enters at the *Initial* orb noting the transformation *power low*, it will execute *seek charge source*. It is the presented with a boundary where it crosses to *discover charging station* or *discover wireless*. The boundary stipulates crossing it in the opposite direction, when the negative exits *lost signal/track* or *no signal* are noted, will place it again at *seek charge source*. After satisfying the condition at *discover wireless*, the robot enters the *find wireless power* state machine at *Initial*. Because its goal is to *find* source, it is presented with two choices—*poll power beacon* and *engage resonance*. It can send out a polling signal to detect feedback whether or not the power signal is in range of its onboard wireless power absorption system. If the *poll power beacon* can lead the robot to the signal, it can then execute *engage resonance*, then exit at *located*.

The wireless power system used in the model relies on the physical principle of coupled modes [19], where the robot has a coil-circuit with selective components that, when engaged, determines the amount of coupling to the power signal [20, 21], translating to the intensity of the power to be absorbed. This activity is represented in Fig.5 by *engage resonance*. In terms similar to the runtime behavior diagram of Fig.4, Fig.6 describes the wireless power scenario including the notion of the domain that encapsulates the different actions in the behavior.

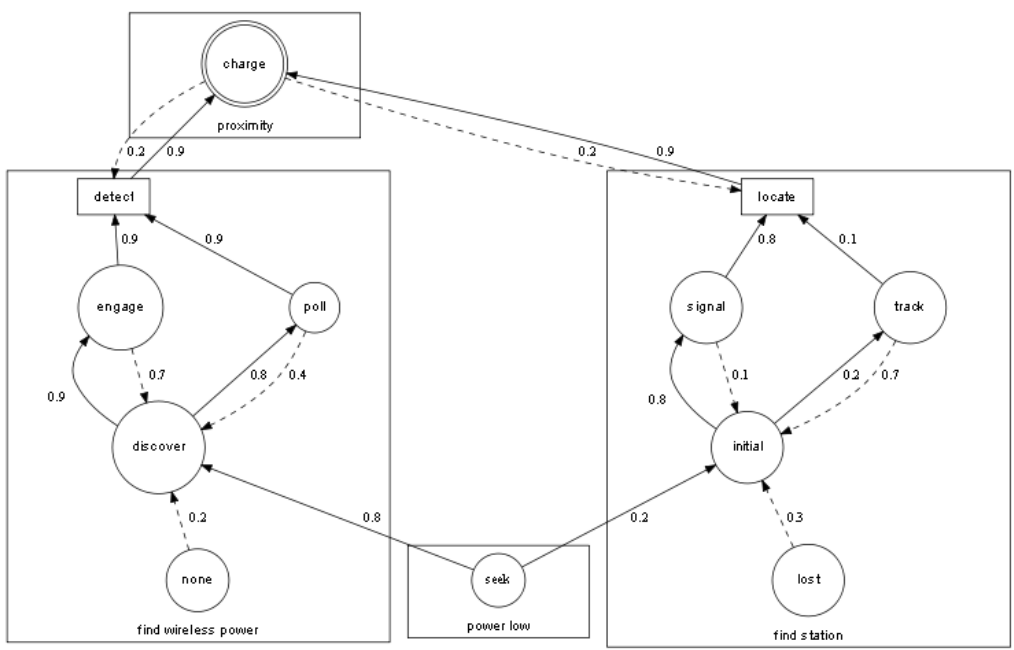

Fig.6. Runtime choice-weights for consequential decision-making in a complex power-signaling scenario.

The robot enters the diagram at the domain *power low* and by reading the weights at *seek*, determines that recharging by using *find wireless power* is better than *find station* given the comparison weights are 0.8 and 0.2, respectively. At the start, *discover* and *initial* also contain negative weights of 0.2 and 0.3, respectively. Following the highest weight into the domain *find wireless power*, it will read at *discover* that *engage* and *poll* have near equal values—within the tolerance of 0.1. However, since the negative weights are 0.7 and 0.4, respectively, the better choice would be to *poll* since it leads to *detect* with a higher probability, a weighted value of 0.9. The transformation between *detect* and *charge* shows the exit from *find wireless power* and entry into *proximity* is strong, with a value of 0.9, with minimal negative feedback, with a value of 0.2.

In this scenario, the weights for each decision path are averaged for each successful result. Having traversed the entirety of choice, or having given the experience by seeding the weight values, the robot knows the optimal path. The ethological implication of the modeled behavior embedded in the diagram of Fig.6 is the additional domain wireless recharging contributes to "laziness". Entry into the domain *find wireless power* will lead to recharging, i.e., reach the goal of *final*, with a high probability of success with the robot having to use the fewest amount of resources to attain its goal of survival. Essentially, the availability of choices to pursue in the model, subtended by weights, allows more degrees of freedom to the runtime yielding insights into how a robot would optimize its choices and how it would express the concept of "laziness" when choosing feeding sources. The array of choice lies between the extremes of charged and uncharged batteries, settling around an equilibrium of constant discharging over time, the weight of which, determined by the amount of runtime activity—searching, processing data, and using sensors. The cooperative nature of the design recalls one final contention: how does the coordination of the activity contribute to the concept of survival?

**The entropic circuit**

Entropy is defined as the number of ways a thermodynamic system can be arranged as it experiences time. A system undergoing entropy would dissociate itself, decaying into disorder. This definition of entropy can be applied to the robotic feeding models in terms of the arrangement of behavior following consequence of choice and the level of sustainability, e.g., the amount of energy the robot possesses at any point during runtime. The measure of entropy would be the success or failure of the robot to attend to its survival, that is, maintain runtime verses shutting down due to the lack of power. Fig.7 depicts a circuit that introduces entropy, a dependency in the onboard energy system wherein the robot is made aware of the ability to control its own survival.

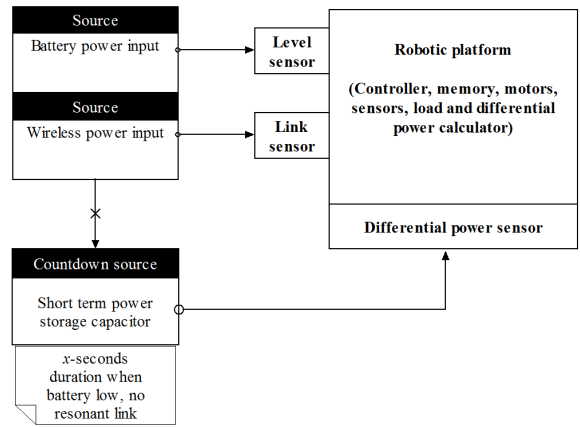

Fig.7. Functional diagram of an entropic circuit.

To the right side of Fig.7 is the robotic platform, this consists of the usual components that a robotic form would take, given its design considerations. To this would be added a sensor to manage the resonant link [19, 20, 21] and a differential power sensor to monitor the value of the operating power and the value of the countdown source. The level sensor measures power stored in the batteries; compared with the value at the differential power sensor, the operating "mood" would be set: normal, seeking, charging, or distressed. To the left side of Fig.7 are the sources of energy—battery and wireless power—and the countdown source which is charged by the wireless power transmission. It contains a short-term storage capacitor of a value indicating the amount of available runtime if the robot had to rely solely on its energy.

Entropy, represented in the functional circuit diagram in Fig.7, is used to model the necessary quantitative information to specify the exact physical state of a system at any given point in time. In other words, the adaption the robot undertakes to counterbalance its effects. Similar to thermodynamic entropy, it is used as a measure of the changes in information manipulation as the runtime evolves from its initial condition at startup. The design allows the freedom for the machine to remain online indefinitely, provided it can execute the set of rules leading to the states for the best outcome. In a theoretical context, the circuit prefers to optimize equilibrium at the expense of decay, analogously to the control of the extremes of life and death by choice and experience by consequence. It reinforces causality. If the robot does not make the right choices to maintain its survival, it turns off and the data in memory is truncated, or lost, depending on the conditions of the experiment. If the robot does not follow results from weights or calls *poll* at a critical moment when it should have called *engage*, it will not receive the proper power. If the experience gained from its life cycle is stored in non-volatile memory, the consequence is written into the program. Otherwise, if volatile memory is used, cessation of activity is the result and the data erased.

**Conclusion**

This paper has described models of robotic feeding and introduced the phenomenon of choice to direct causal outcomes following consequence. It has discussed two detailed models of different feeding strategies, which have yielded an experimental paradigm to empirically test the fuzzy concept of choice, survival, and the consequence of behavior following knowledge of entropy. The models presented here yield a richer set of results and give a means to validate whether or not artificial systems possesses what biological systems classify as the survival instinct. Some of the instincts to manifest are: where the robot has problems accessing the station, by mechanical error or environmental issues, coupling to the energy field, or that it commits suicide by giving up to the imposed condition of the entropic circuit. Decision-making is internalized by the robot and removed from an interaction with humans. The system is independent and free from external intervention apart from initial conditioning or implementation of the starter program that will provide the machine its beginning point and tasks to perform during its life cycle including sample weights to preclude consequential decisions.

A new and distinct methodology offers not only model and quantify artificial life but to illustrate a method whereby a machine can construct an understanding about unknown events, while arriving at insights about the structure of the environment it finds itself in. The power in this approach is the artificial system more closely mimics a living system by collecting memory of each experience navigating the finite-states of feeding behavior. To facilitate this, the composite state machine presented here is contained in a modular, domain-specific language. It describes a set of features relevant to the space in which the program will operate which allows the creation of members aiding in a more robust development methodology [22]. This is important in this context because, as in the case with Walter, a rigidly defined platform is a mathematical framework yielding higher orders of behavior, the quantification of which reduces the need for speculation by the observer. Theoretically, a machine would come to realize that it has the power to control its own life and death.

The study of robotic feeding models, choice, and the survival mechanism contributes to the body of knowledge surrounding research into the properties of cognition in a very succinct way. Rodney Brooks notes [23]:

> Researchers in artificial intelligence and artificial life are interested in understanding the properties of living organisms so that they can build artificial systems that exhibit these properties for useful purposes. AI researchers are interested mostly in perception, cognition and generation of action, whereas artificial life focuses on evolution, reproduction, morphogenesis and metabolism. Neither of these disciplines is a conventional science; rather, they are a mixture of science and engineering. Despite, or perhaps because of, this hybrid structure, both disciplines have been very successful and our world is full of their products.

The advantage for an entropic circuit lies in a continuous set of activities for the robot and observation of long-term behavior. It will lead to advances in understanding artificial ethological concepts executing routines in the scope of the domains. Additionally, as a composite phenomenon, it can replete an ordinary robot with artificial life. Furthermore, a global picture of behavior of the sum of transformations between states and the choices can be used to describe the "personality" of the robot's runtime. The pursuit of equilibrium between the extremes of a requisite variety of the choice between different recharging models—a station or a wireless power source—follows closely the description first set out by Walter.

**References**

[1] W.G. Walter, "An electromechanical animal," *Dialectica*, Vol. 4, pp. 42-9, 1950.

[2] O. Holland, "Grey Walter: The pioneer of real artificial life," *Proceedings of the 5th International Workshop on Artificial Life*, MIT Press, Cambridge, pp. 34-44, 1997.

[3] L. Smith and C. Breazeal. "The dynamic life of developmental process," *Developmental Science*, (1), 61-68, 2007.

[4] R. Brooks, "Intelligence without representation," *Artificial Intelligence*, 47 (1-3): pp. 139–59, doi:10.1016/0004-3702(91)90053-M.

[5] W.G. Walter, "An imitation of life," *Scientific American*, 182(5): 42-5, 1950.

[6] W.G. Walter, "A machine that learns," *Scientific American*, 185(2): 60-3, 1951.

[7] O. Holland and R. Goodman. "Robots with internal models: a route to machine consciousness?" *Journal of Consciousness Studies, Special Issue on Machine Consciousness*, Vol. 10, No. 4, 2003.

[8] S. Coradeschi, H. Ishiguro, M. Asada, S. Shapiro, M. Theilscher, C. Breazeal, M. Mataric, and H. Ishida. "Human-Inspired Robots," *IEEE Intelligent Systems*, 21(4), 74-85, 2006.

[9] O. Holland, "Exploration and high adventure: The legacy of Grey Walter," *The Royal Society*, Aug. 2003.

[10] W.G. Walter, *The Living Brain*, New York, 1953.

[11] G. Schönera, M. Doseb, and C. Engelsb, "Dynamics of behavior: Theory and applications for autonomous robot architectures," *Robotics and Autonomous Systems*, Vol. 16, Issues 2–4, pp. 213–45, Dec. 1995.

[12] M. Proetzsch, T. Luksch, and K. Berns, "Development of complex robotic systems using the behavior-based control architecture iB2C," *Robotics and Autonomous Systems*, Vol. 58, Iss. 1, pp. 46–67, Jan. 2010.

[13] R. Brooks, "A robust layered control system for a mobile robot," *IEEE Robotics and Automation*, Vol. 2 Iss. 1, Sep. 1985.

[14] J. Košeckà and R. Bajcsy, "Discrete event systems for autonomous mobile agents," *Robotics and Autonomous Systems*, Vol. 12, Issues 3–4, pp. 187–98, Apr. 1994.

[15] M.J. Matarić, "Learning social behavior," *Robotics and Autonomous Systems*, Vol. 20, No. 2, pp. 191–204, Jun. 1997.

[16] W.R. Ashby, *Introduction to Cybernetics*, London, Chapman & Hall, Second Impression, 1957.

[17] W.R. Ashby, *Design for a Brain*, New York, Wiley, 1954.

[18] C.R. Melhuish, O. Holland, and S.E.J. Hoddell. "Using chorusing for the formation of travelling groups of minimal agents," *Robotics and Autonomous Systems*, Vol. 28, pp. 207-16, 1999.

[19] C.A. Tucker, "Magnetic resonant modes in a wireless-powered circuit," *19$^{th}$ IEEE Telecommunications Forum (TELFOR)*, pp. 977-80, Nov. 2011.

[20] C.A. Tucker, K. Warwick, and W. Holderbaum, "A contribution to the wireless transmission of power," *IJ of Electrical Power and Energy Systems*, Vol. 47, pp. 235-42, May 2013.

[21] C.A. Tucker, "System of transmission of wireless power," U.S. Patent 8,274,178, Sep. 2012.

[22] A. Van Deursen and P. Klint. "Domain-specific language design requires feature descriptions," *CIT. Journal of computing and information technology*, 10.1, pp. 1-17, 2002.

[23] R. Brooks, "The relationship between matter and life," *Nature*. Vol. 409, No. 6818, pp. 409-11, 2001.